\documentclass[manuscript,screen]{acmart}
\usepackage{tabularx}
\usepackage{soul}
\usepackage{url}
\usepackage{graphicx}
\usepackage{float}
\usepackage{amsmath}
\usepackage{booktabs}
\usepackage{subfigure}
\usepackage{algorithm}
\usepackage{algorithmic}
\usepackage{amsmath}
\usepackage{multirow}
\usepackage{comment}
\usepackage{makecell}
\usepackage{bm}
\usepackage{color}
\AtBeginDocument{%
  }

\setcopyright{acmcopyright}
\copyrightyear{2018}
\acmYear{2018}
\acmDOI{XXXXXXX.XXXXXXX}

\acmConference[Conference acronym 'XX]{Make sure to enter the correct
  conference title from your rights confirmation emai}{June 03--05,
  2018}{Woodstock, NY}
\acmPrice{15.00}
\acmISBN{978-1-4503-XXXX-X/18/06}

\begin{document}

\title{Multi-view Graph Convolutional Networks with Differentiable Node Selection}

\author{Zhaoliang~Chen}
\email{chenzl23@outlook.com}
\orcid{0000-0002-7832-908X}
\email{webmaster@marysville-ohio.com}
\affiliation{%
  \institution{College of Computer and Data Science, Fuzhou University}
  \city{Fuzhou}
  \country{China}
  \postcode{350116}
}

\author{Lele Fu}
\affiliation{%
  \institution{School of Systems Science and Engineering, Sun Yat-sen University}
  \city{Guangzhou}
  \country{China}
  \postcode{350116}
\email{lawrencefzu@gmail.com}
}

\author{Shunxin Xiao}
\affiliation{%
  \institution{College of Computer and Data Science, Fuzhou University}
  \city{Fuzhou}
  \country{China}
  \postcode{350116}
\email{xiaoshunxin.tj@gmail.com}
}

\author{Shiping Wang}
\affiliation{%
  \institution{College of Computer and Data Science, Fuzhou University}
  \city{Fuzhou}
  \country{China}
  \postcode{350116}
\email{shipingwangphd@163.com}
}

\author{Claudia Plant}
\affiliation{%
  \institution{Faculty of Computer Science, University of Vienna}
  \city{Vienna}
  \country{Austria}
  \postcode{1090}
\email{claudia.plant@univie.ac.at}
}

\author{Wenzhong Guo}
\authornote{The corresponding author.}
\affiliation{%
  \institution{College of Computer and Data Science, Fuzhou University}
  \city{Fuzhou}
  \country{China}
  \postcode{350116}
\email{guowenzhong@fzu.edu.cn}
}

\renewcommand{\shortauthors}{Chen et al.}

\begin{abstract}
  Multi-view data containing complementary and consensus information can facilitate representation learning by exploiting the intact integration of multi-view features.
  Because most objects in real world often have underlying connections, organizing multi-view data as heterogeneous graphs is beneficial to extracting latent information among different objects.
  Due to the powerful capability to gather information of neighborhood nodes, in this paper, we apply Graph Convolutional Network (GCN) to cope with heterogeneous-graph data originating from multi-view data, which is still under-explored in the field of GCN.
  In order to improve the quality of network topology and alleviate the interference of noises yielded by graph fusion, some methods undertake sorting operations before the graph convolution procedure.
  These GCN-based methods generally sort and select the most confident neighborhood nodes for each vertex, such as picking the top-$k$ nodes according to pre-defined confidence values.
  Nonetheless, this is problematic due to the non-differentiable sorting operators and inflexible graph embedding learning, which may result in blocked gradient computations and undesired performance.
  To cope with these issues, we propose a joint framework dubbed Multi-view Graph Convolutional Network with Differentiable Node Selection (MGCN-DNS), which is constituted of an adaptive graph fusion layer, a graph learning module and a differentiable node selection schema.
  MGCN-DNS accepts multi-channel graph-structural data as inputs and aims to learn more robust graph fusion through a differentiable neural network.
  The effectiveness of the proposed method is verified by rigorous comparisons with considerable state-of-the-art approaches in terms of multi-view semi-supervised classification tasks,
  and the experimental results indicate that MGCN-DNS achieves pleasurable performance on several benchmark multi-view datasets.
\end{abstract}



\received{20 February 2007}
\received[revised]{12 March 2009}
\received[accepted]{5 June 2009}

\maketitle

\section{Introduction}
Multi-view data extensively exist in a great number of practical applications, owing to the fact that most objects in the real world can be described from multifarious perspectives \cite{10.1145/3532191,ChenWPHZ21,LiuMulti20,lu2022differentiable}.
A large amount of research has verified that multi-view data can lead to encouraging performance compared with single-view data, such as the semi-supervised classification task that only has scarce supervision information.
Because objects in real world often lead to latent connections and can be regarded as nodes, we can construct heterogeneous graphs from multi-view data which depict objects more comprehensively.
Accordingly, graph-based learning models can be adapted to multi-view learning tasks.
Graph learning is a critical technique for many machine learning tasks like clustering and classification \cite{10.1145/3564701,LiHWHLC21,TangZLLWZW19,YanhuiHybrid2022}.
In particular, Graph Convolutional Network (GCN) is a crucial technique of graph-based representation learning and has been universally applied to a multitude of machine learning tasks.
GCN-based methods have been well-established in computer vision \cite{XieLSC20,liu2021dgig,YangYZSLM22}, classification tasks \cite{zhang2021shne,JiaLWZNHZ20,DongLDZ22} and topological graph structure analyses \cite{chen2021dacha,xia20213dgcn,wei2022dual}.
In a nutshell, graph convolutions are performed in the non-Euclidean domain instead of extensively employed standard convolutions for the Euclidean domain, which have a powerful ability to integrate the connectivity patterns and feature attributions of graphs in a Fourier space.
Actually, GCN is a typical network that aggregates topological neighbors via convolutional layers, so that node and feature representations propagate over the network topology.
Attributed to these factors, frameworks based on GCN have achieved noticeable performance for extracting robust numeric representations.
However, a traditional GCN-based method only enables single-view data, meaning that further study on frameworks addressing multi-graph-structural data should be conducted.
Some existing methods have endeavored to tackle this deficiency with a linear weighted sum of heterogeneous graphs \cite{LiLW20a,ZhangQWN21}, but it may not be ideal because such a strategy can not extract deeper representation and results may be interfered by new noises encountered.
Consequently, a study on multi-channel GCN-based models learning more robust and generalized topological graphs should be further explored.

\begin{figure}[!htbp]
  \centering
  \includegraphics[width=0.8\textwidth]{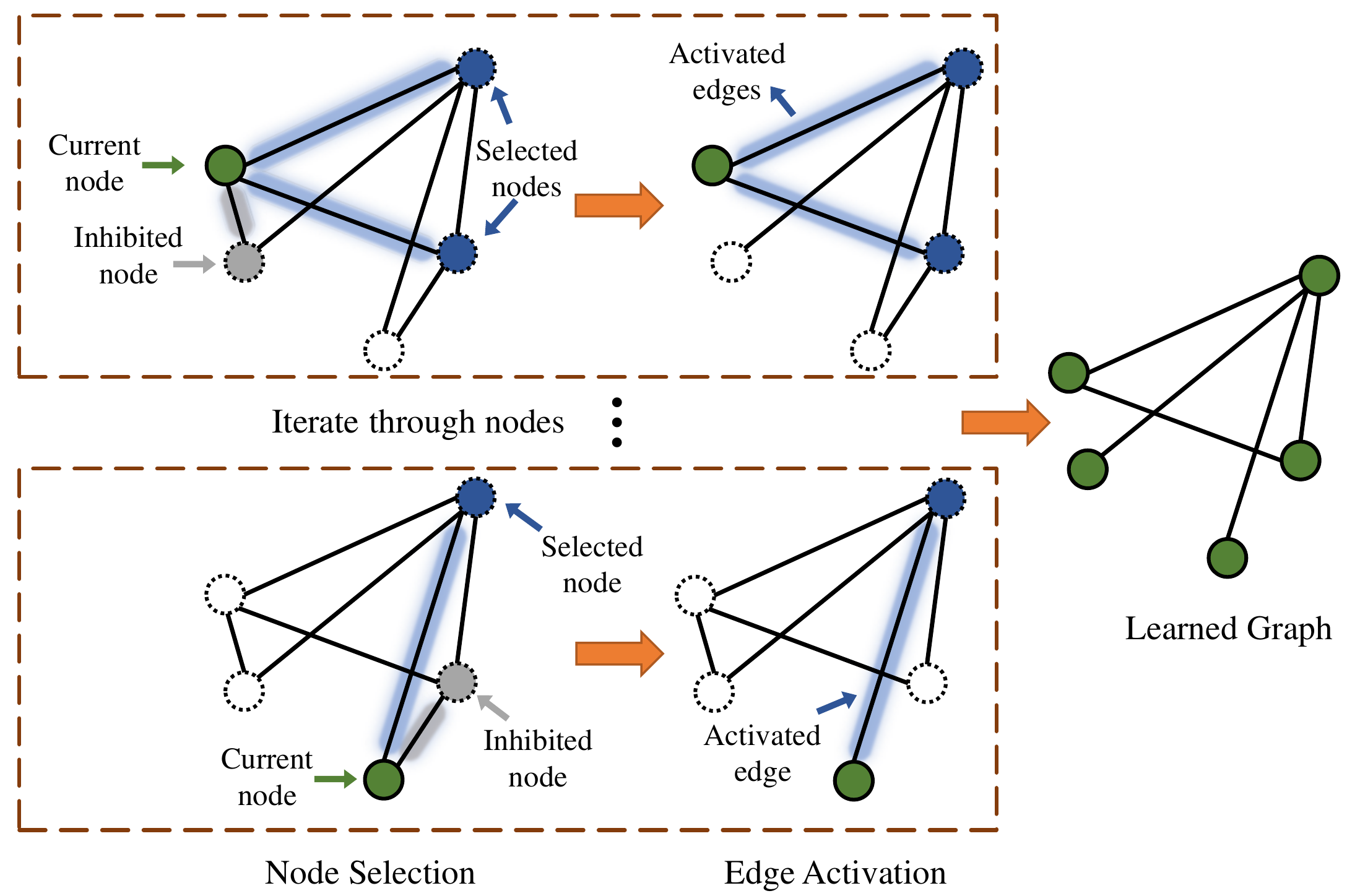}
  \caption{Node selection for a graph-structural network.
  For each node (green), only other connected nodes with higher importance or confidence (blue) are selected.
  Namely, salient edges (blue) connecting to important nodes are activated and other edges (gray) are inhibited.
  Thus, a sparser graph is learned due to the restraint of some edges.
  }
  \label{NodeSelection}
\end{figure}

Node sorting and selection strategies have been utilized in numerous works related to GCN to pick top-$k$ confident nodes, which prompt the sparsity and lessen the interference of noises \cite{WuPL0CZ20,hui2020collaborative,SunLZ20}.
In addition, it is helpful for sustained explorations of relationships between every two vertices and augmenting the edges with higher importance, because of which the time consumption declines and the performance is improved.
Figure \ref{NodeSelection} briefly illustrates the node selection procedure, which selects neighborhood nodes with high confidence or importance for each vertex.
The confidence or importance values can be measured by the weights of edges or feedback from downstream tasks.
Nevertheless, such a strategy is generally non-differentiable, thereby often leading to ineffective gradient propagation, which is problematic for neural networks.
Consequently, it becomes pivotal to delve into a differentiable network allowing gradient descent and back propagation everywhere.
Although Grover et al. presented a differentiable operator generating permutation matrix \cite{GroverWZE19}, the sorting and selection operations are still performed with $\arg \max$ operator which is non-differentiable and prohibits the gradient propagation.
Therefore, we devote ourselves to establishing a differentiable learning schema that automatically selects confident nodes and maintains the differentiability of the neural network.

\begin{figure*}[!htbp]
  \centering
  \includegraphics[width=\textwidth]{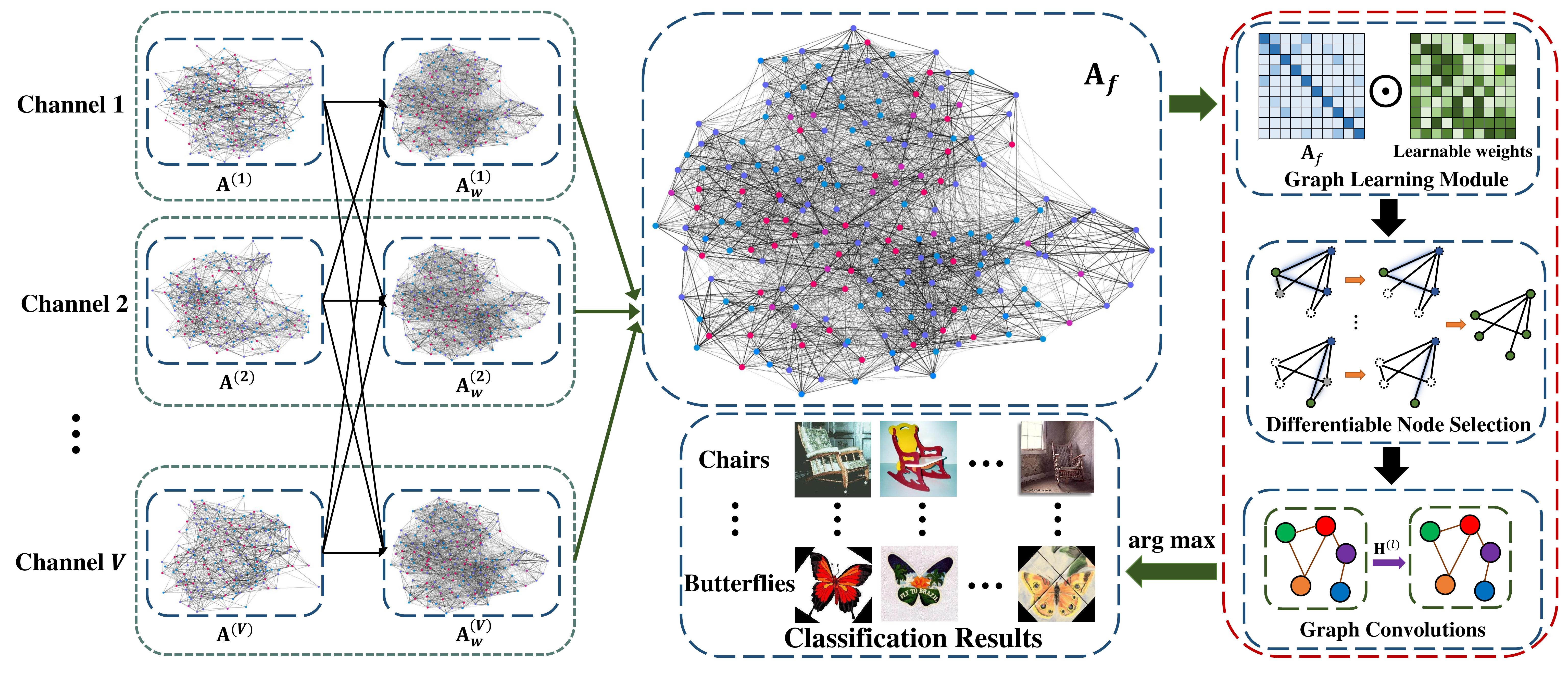}
  \caption{Structure of the proposed MGCN-DNS, which endeavors to solve multi-view semi-supervised classification problems.
  After the two-stage adaptive fusion of all graphs, the learned intact adjacency matrix is dexterously manipulated by the graph learning module and differentiable node selection process, respectively. Eventually, the refined adjacency matrix is exploited in graph convolutional networks to produce node embeddings that indicate the classification results.
  }
  \label{Framework}
\end{figure*}

In pursuit of solving these mentioned issues, this paper proposes a GCN-based framework dubbed Multi-view Graph Convolutional Network with Differentiable Node Selection (MGCN-DNS) to undertake multi-view semi-supervised classification tasks. 
As shown in Figure \ref{Framework}, the proposed method addresses multi-graph-structural data via a graph fusion layer and extracts numerical representations of node relationships through a graph learning module and a novel differentiable node selection schema.
In detail, adjacency matrices from heterogeneous views serve as the inputs for each channel in the graph fusion layer, which explores the underlying complementary information from all channels and outputs the combined adjacency matrix.
Next, the intact graph is refined by a graph learning module and a  differentiable node selection module for each vertex.
To our knowledge, this work is the first attempt to integrate differentiable node selection schema into the adjacency matrix learning of GCN-based models.
Finally, the refined adjacency matrix is employed by graph convolutions to yield node embeddings.
In a nutshell, the main contributions of this paper are outlined as follows:

(1) Devise a two-stage graph fusion process that exploits the complementary information from adjacency matrices in various channels via trainable weights.

(2) Develop a differentiable node selection procedure to choose nodes with higher confidence, which converts the continuous probabilistic permutation matrix to confidence values measuring the importance of distinct nodes.

(3) Construct an end-to-end graph convolutional network to deal with multi-channel topological graphs, which extracts neighborhood relationships via graph learning modules and differentiable node selection.

(4) The proposed framework is applied to undertaking the multi-view semi-supervised classification task and achieves pleasurable performance in comparison with classical and state-of-the-art methods.

The rest of this paper is organized as follows.
Related works including multi-view learning, multi-graph learning and GCN are discussed in Section \ref{Relatedworks}.
We elaborate on the proposed MGCN-DNS and relevant theoretical analysis in Section \ref{framework}.
In Section \ref{experiments}, substantial experiments and comparisons are conducted to show the superiority of the model.
Finally, we conclude our work and discuss future directions in Section \ref{conclusion}.

\section{Related Work}\label{Relatedworks}
In this section, we discuss some relevant works w.r.t. multi-view learning and GCN.
Before that, the primary notations and explanation of mathematical symbols used in this paper are clarified in Table \ref{Notations}.

\begin{table*}[!htbp]
  \center
  \begin{tabular}{l|l}
  \toprule
   Notations&       Explanations   \\
   \midrule
   $\mathbf{X}^{(v)}$  &  Feature matrix of the $v$-th view. \\
  $\mathbf{A}^{(v)}$  &  Adjacency matrix of the $v$-th view.\\
   $w^{(i,j)}$ &  Adaptive weight of the adjacency matrix.  \\
  $\mathbf{A}_{w}^{(v)}$  & Complementary graph of the $v$-th view. \\
   $\alpha^{(i)}$ &  Weight of the $i$-th complementary graph. \\
     $\mathbf{A}_{f}$ & Fused intact adjacency matrix. \\
   $\hat{\mathbf{A}}$  & Refined graph via graph learning module.  \\
     $\mathbf{S}_{1}, \mathbf{S}_{2}$ & Weight matrices of graph learning module. \\
   $\mathbf{P}$ &  Continuous permutation matrix. \\
     $\text{Softmax}(\cdot)$ &  Softmax function. \\
   $\gamma$  & Hyperparameter in graph learning module. \\
     $\tau$  &  Temperature parameter in the construction of $\mathbf{P}$. \\
   $\mathcal{I}_{i}$ &  Node confidence value.      \\
     $\mathbf{C}$ &  Coefficient matrix for node selection. \\
   $\theta$ &  Learnable thresholder.  \\
    $\mathbf{A}_{select}$ &  Adjacency matrix after differentiable node selection. \\
   $c$ &  Number of classes.  \\
    $\sigma(\cdot)$  & Optional activation function.   \\   
   $\mathbf{A}_{s}$   & Adaptively learned adjacency matrix fusion. \\
      $\mathbf{W}^{(l)}$       & Trainable weight matrix in graph convolutions.  \\
   $\mathbf{H}^{(l)}$        & Output of GCN in the $l$-th layer. \\
     $\mathbf{Z}$         &  Learned graph embedding.  \\
  $\mathbf{Y}$  & Groundtruth of labeled samples.  \\
    $\mathcal{L}$  & Loss function. \\
  \bottomrule
  \end{tabular}
  \caption{A summary of primary notations in this paper.}
  \label{Notations}
\end{table*}

\subsection{Multi-view Learning}
In real-world practical applications, a considerable number of data exist as heterogeneous forms, leading to a rapidly increasing attention on multi-modality fusion.
Consequently, multi-view learning has been extensively studied in recent years.
Liu et al. improved the performance of multi-view clustering by considering information from multiple incomplete views. \cite{LiuLTXXLKZ21}.
Chen et al. proposed a partial tubal nuclear norm regularized multi-view learning algorithm, which minimized the partial accumulation of the smaller tubal singular values to keep the low-rank property of the self-representation tensor \cite{ChenWPLZ21}.
A semi-supervised structured subspace clustering model was proposed for multi-view data, which explored the block-diagonal architecture of the shared affinity matrix \cite{QinWZF22}.
Han et al. performed multi-view clustering via establishing a sparse membership matrix over heterogeneous views and studied the centroid matrix and its corresponding weight of each view \cite{HanXNL22}.
Wang et al. put forward a parameter-free multi-view subspace clustering model with consensus anchor guidance, which had linear computational complexity \cite{WangLZZZGZ22}.
Tao et al. employed $\ell_{2,1}$-norm to compute the regression losses of heterogeneous views, which constructed the objective function with the weighted accumulation of all regression losses \cite{TaoHNZY17}. 
Yang et al. proposed a multi-view classification algorithm where features from assorted views were adaptively assigned with the learned optimal weights \cite{YangDN19}.
Huang et al. established a framework integrating diversity, sparsity and consensus to flexibly process multi-view data, which projected a linear regression model onto deduce view-specific embedding regularizers and automatically updates weights of various perspectives \cite{HuangWZZL21}.
Han et al. explored multi-human association on multi-view images taken by various cameras at the same moment \cite{HanWYFW22}.
Yan et al. proposed a bipartite graph-based discriminative feature learning algorithm to perform multi-view clustering, which was a unified framework to consider bipartite graph learning and discriminative feature learning \cite{YanXLYT22}.
Zhang et al. constructed anchor-based affinity graphs to address the multi-view semi-supervised problem \cite{ZhangQWN22}.

\subsection{Multi-graph Fusion}

Owing to the fact that most multi-view data can be represented as feature attributes and node relationships incorporated in distinct network topologies, multi-graph fusion has become a popular issue.
Some of these methods merged graph information from natural graph data.
Sadikaj et al. designed a SpectralMix algorithm which integrated all information
from different node attributes, relationships and graph structures to cope with multi-relational graphs \cite{SadikajVBP21}.
Tang et al. built a multi-graph-fusion-based spectral clustering method via proposing a hyperspectral band selection algorithm \cite{TangLZ0Z21}.
A multi-graph network was presented to cope with unsupervised 3D shape retrieval, employing correlations among modalities and structural similarity between every two models \cite{NieZLGS20}.
Sun et al. put forward multi-view GCN to estimate the crowd flow, which fused the outputs of view-specific GCNs \cite{sun2020predicting}.
A self-supervised framework was constructed by Hassani et al. to learn node and graph level representations simultaneously \cite{HassaniA20}.
Some other works also integrated heterogeneous graphs generated from multi-view data which are more common in real-world applications.
For example, a co-training strategy was exploited to explore a unified GCN-based framework, where the graph information embedded in sundry views was trained automatically \cite{LiLW20a}.
Graphs of multiple views were incorporated into a consistent global graph, whose Laplacian matrix was subjected to multiple strongly connected components \cite{NieLiLi17SelfWeighted}.
Nie et al. designed a parameter-free framework that automatically assigned weights to each view and explored a label indicator matrix by collecting loss functions of all views \cite{NieLiLi16Parameter}.
Nie et al. proposed a local structure embedding model that integrated unsupervised clustering and semi-supervised classification into a unified framework \cite{NieCaiLi17Multiview}.
Xie et al. constructed a unified tensor space to jointly explore multi-view correlations by local geometrical structures, where a low-rank tensor regularization was imposed to guarantee that all views come to an agreement \cite{XieZQDT20}.
Wang et al. learned sparse graphs from heterogeneous views via a deep sparse regularization network to conduct multi-view semi-supervised classification and clustering \cite{WangCDL22}.
Most of these prior works concentrated on exploring an effective paradigm to merge graphs with linear combinations or attention mechanisms.
In this paper, we further build a two-stage graph fusion operation for all initial adjacency matrices, which merges robust neighborhood correlations from all views and explores complementary information via trainable weights.

\subsection{Graph Convolutional Network}
A spectral graph convolution is performed by a signal $x \in \mathbb{R}^{m}$ and a filter $g_{\theta} = diag(\theta)$, represented as 
\begin{gather}\label{OriginalGCN}
g_{\theta} \star x=\mathbf{U} g_{\theta} \mathbf{U}^{\top} x,
\end{gather}
where $\mathbf{U}$ is the matrix of eigenvalues.
Kipf et al. \cite{KipfW17} conducted the first-order approximation of truncated Chebyshev polynomial to simplify the spectral graph convolution, and applied it to the semi-supervised classification, where data were organized as network topology.
In particular, a convolutional layer of GCN is formally defined as
\begin{gather}\label{GCN}
\begin{split}
\mathbf{H}^{(l)} = \sigma \left( \tilde{\mathbf{D}}^{-\frac{1}{2}} \tilde{\mathbf{A}} \tilde{\mathbf{D}}^{-\frac{1}{2}} \mathbf{H}^{(l-1)} \mathbf{W}^{(l)} \right),
\end{split}
\end{gather}
where $\tilde{\mathbf{A}} = \mathbf{A} + \mathbf{I}$ is the adjacency matrix considering the self-connections, $\mathbf{H}^{(l)}$ is the learned graph embedding in the $l$-th layer, and $[\tilde{\mathbf{D}}]_{ii} = \sum_{j} [\tilde{\mathbf{A}}]_{ij}$.
To simplify notations, in this paper, we directly replace the renormalized $\tilde{\mathbf{D}}^{-\frac{1}{2}} \tilde{\mathbf{A}} \tilde{\mathbf{D}}^{-\frac{1}{2}}$ with $\mathbf{A}$.
Numerous studies have paid attention to GCN-based frameworks, exploiting the capability of gathering node attributes to promote the performance of varying machine learning tasks.
Zhu et al. proposed a variant of GCN derived via a modified Markov diffusion kernel, which explored the global and local contexts of nodes \cite{ZhuK21}.
Ye et al. established a context-encoding network with GCN to study skeleton topology adaptively \cite{YePZLXT20}.
Liu et al. integrated GCN with hidden conditional random fields to reserve the skeleton structure information of people during the classification stage \cite{LiuGKQG21}.
Wang et al. mined the specific and common embeddings from node features, topological structures, and their combinations simultaneously via a multi-channel GCN, whose adaptive importance weights were trained via the attention mechanism \cite{0017ZB0SP20}.

For the purpose of promoting the sparsity and performance of graph-based models, a population of methods has attempted to select connective nodes with higher confidence for each vertex.
Wu et al. chose the top-$k$ largest values of the adjacency matrix to explore more robust node relationships and improve the accuracy of traffic flow estimation \cite{WuPL0CZ20}.
Nodes with low confidence values were randomly selected to build subgraphs for the GCN framework \cite{zhang2020global}.
Hui et al. selected labeled nodes with top-$k$ confidence scores during training iterations \cite{hui2020collaborative}.
Sun et al. sorted the confidence values of vertices to improve the quality of graphs in GCN-based methods \cite{SunLZ20}.
In light of these prior works, node selection is generally conducted by a sorting operator or top-$k$ operator denoted by $\arg topk$.
Algorithm \ref{AlgorithmKNN} illustrates a common node selection process with top-$k$ operator in most GCN-based frameworks.
These methods generally preserved the top-$k$ most important or confident neighborhood nodes of each vertex, according to the weights of edges,
which resulted in a sparser adjacency matrix after the node selection operation.
Nevertheless, these models face a problematic issue that they are not differentiable everywhere owing to the non-differentiable node sorting and selection procedure, which may result in the gradient vanishing and blocking of gradient propagation during training.
For the sake of tackling this problem, in this paper, we construct a differentiable learning progress that figures out the confidence score of each node and automatically selects vertices with higher confidence.

\begin{algorithm}[!tbp]
  \renewcommand{\algorithmicrequire}{\textbf{Input:}}
  \renewcommand{\algorithmicensure}{\textbf{Output:}}
  \caption{A Common Algorithm for Node Selection }
  \begin{algorithmic}\label{AlgorithmKNN}
  \REQUIRE Adjacency matrix $\mathbf{A} \in \mathbb{R}^{m \times m}$.
  \ENSURE Adjacency matrix $\mathbf{A}_{select}$ after node selection.
  \STATE {Initialize $k$ of top-$k$ ranking;}
  \STATE {Initialize $\mathbf{A}_{select}$ as a null matrix;}
      \FOR {$i = 1 \rightarrow m$}
          \STATE {Generate the node connectivity patterns of the $i$-th node by $\mathbf{a}_{i} = [a_{i1}, \cdots, a_{im}]$, where $a_{ij} = [\mathbf{A}]_{ij}$;}
          \STATE {Find the index of the top-$k$ largest (or the most important) nodes by $idx = \arg topk (\mathbf{a}_{i})$;}
          \STATE {Conduct $\mathbf{A}_{select}[i,idx] = \mathbf{a}_{i}[idx]$;}
      \ENDFOR
  \RETURN {Selected adjacency matrix $\mathbf{A}_{select}$.}
  \end{algorithmic}
  \end{algorithm}

\section{Multi-view Graph Convolutional Network with Differentiable Node Selection}\label{framework}
Multi-view semi-supervised classification is a widely investigated problem in recent years, which conduct classification tasks with very limited labeled samples.
It aims to overcome the scarce semi-supervised signals via making full use of information from heterogeneous views. 
Given multi-view data $\mathcal{X} = \{\mathbf{X}^{(v)} \}_{v=1}^{V}$,
where $\mathbf{X}^{(v)} \in \mathbb{R}^{m \times n}$ represents the feature matrix of the $v$-th view with totally $V$ views,
the proposed MGCN-DNS is supposed to tackle the multi-view semi-supervised classification problems with renormalized adjacency matrices $\mathcal{A} = \{\mathbf{A}^{(v)} \}_{v=1}^{V}$ generated from $\mathcal{X}$, where $\mathbf{A}^{(v)} \in \mathbb{R}^{m \times m}$.
Adjacency matrices from distinct channels exploit complementary information by a two-stage adaptive graph fusion layer, after which MGCN-DNS refines the fused adjacency matrix via a graph learning module and a differentiable node selection procedure.
Finally, the refined intact adjacency matrix is applied to conducting graph convolution operations to yield node embeddings.

\subsection{Multi-channel Graph Fusion}\label{MGF}
First of all, the adjacency matrices are initialized via the $k$-Nearest Neighbors (KNN) method.
We first perform an adaptive weighted sum of initial adjacency matrices from all views to leverage complementary information from other perspectives. Namely,
\begin{gather}\label{WeightedA}
\begin{split}
\mathbf{\mathbf{A}}_{w}^{(v)} = \sum_{i=1}^{V} w^{(v,i)} \mathbf{A}^{(i)},
\end{split}
\end{gather}
where $w^{(v,i)}$ is the trainable weight which indicates the importance of the $i$-th view graph to the $v$-th view graph.
Graph $\mathbf{A}_{w}^{(v)}$ of the $v$-th view can be regarded as a complementary graph because it is enriched by information from other views.
We also utilize the $\text{softmax}$ renormalization defined in Equation \eqref{NormalizationPai} for each channel
\begin{gather}\label{NormalizationPai}
\begin{split}
w^{(v, i)} \leftarrow \frac{\mathrm{exp} \left( w^{(v, i)} \right)}{ \sum_{i=1}^{V}  \mathrm{exp} \left(w^{(v, i)} \right) }
\end{split}
\end{gather}
to guarantee that $\sum_{i=1}^{V} w^{(v, i)} = 1$ during training.
Consequently, a larger $w^{(v, i)}$ trained automatically indicates that the corresponding adjacency matrix in the $i$-th view should make up more for missing information in the $v$-th view.
In other words, if $\sum_{v=1}^{V} w^{(v, i)}$ is larger, the graph in the $i$-th view should be more important because it provides more complementary information to all views.
Therefore, we compute $\alpha^{(i)}$ measuring the contribution of the $i$-th view to all views via
\begin{equation}\label{ComputeAlpha}
  \alpha ^{(i)} = \sum_{v=1}^{V} w^{(v, i)},
\end{equation}
which is renormalized by
\begin{equation}\label{NormalizationAlpha}
    \alpha ^{(i)} \leftarrow \frac{\alpha ^{(i)}}{\sum_{i=1}^{V} \alpha ^{(i)}}.
\end{equation}
Because $w^{(v, i)}$ measures how much additional information is provided by the $i$-th view, an adjacency matrix with a higher value of corresponding $\alpha$ should contribute more to the final fused node relationship information,
thereby containing more complementary information.
Actually, if the value of $\alpha$ is tiny, it indicates that the adjacency in this view has little complementary information and is redundant for the other views.
Thus, we should consider reducing the impact of the corresponding adjacency matrix.
In light of the previous analysis, the unique merged adjacency matrix is obtained by considering this factor:
\begin{gather}\label{FusedA}
  \begin{split}
  \mathbf{\mathbf{A}}_{f} = \sum_{i=1}^{V} \alpha ^{(i)} \mathbf{A}_{w}^{(i)}.
  \end{split}
\end{gather}

\subsection{Graph Learning Module}\label{MGF}
After combining network typologies via the two-stage graph fusion, we construct a graph learning module to further refine the integrated adjacency matrix with the given downstream task.
Because GCN assumes that the graph information should be symmetric and bi-directional,
the outputs of the learnable layer also ought to hold this assumption.
Denoting learnable weight matrices as $\mathbf{S}_{1}, \mathbf{S}_{2} \in \mathbb{R}^{m \times m}$,
the graph learning module is formulated as
\begin{equation}\label{LearnS}
  \hat{\mathbf{A}} =\mathbf{A}_{f} \odot \text{Sigmoid} \left( \gamma \left| \mathbf{S}_{1} {\mathbf{S}_{2}}^{T} - \mathbf{S}_{2}{\mathbf{S}_{1}}^{T} \right| \right),
\end{equation}
where $\odot$ is the Hadamard product and $\mathbf{S}_{1}, \mathbf{S}_{2}$ are randomly initialized.
Hyperparameter $\gamma$ controls the saturation rate of the activation function.
Thus, the learned $\hat{\mathbf{A}}$ keeps properties that edges of topological graphs should be non-negative and symmetric,
which is achieved by trainable matrix $\text{Sigmoid} \left( \gamma \left| \mathbf{S}_{1} {\mathbf{S}_{2}}^{T} - \mathbf{S}_{2}{\mathbf{S}_{1}}^{T} \right| \right)$ serving as an adaptive shrinkage weight matrix.
Herein, $\text{Sigmoid} (\cdot)$ performs the projection to make sure that all entries range in $(0,1)$, which dexterously avoids excessive values and inhibits unimportant connections in a data-driven way during training.
Therefore, an adjacency matrix is refined via learnable weights $\mathbf{S}_{1}, \mathbf{S}_{2}$ in the graph learning module.
Because the trainable weights $\mathbf{S}_{1}, \mathbf{S}_{2}$ are automatically  optimized according to the cross entropy loss defined in \eqref{GCNLoss},
it approximates the optimal graph tailored for the downstream tasks at the best.

\subsection{Differentiable Node Selection}\label{DNS}
So as to further alleviate the influence of noises and make full use of pivotal neighborhood relationships, the proposed method conducts differentiable node selection in each channel to pick essential nodes and promote the sparsity of graphs, on the basis of the probabilistic permutation matrix $\mathbf{P} = [\mathbf{p}_{1}, \cdots, \mathbf{p}_{m}]^{T}  \in \mathbb{R}^{m \times m} $.
In particular, $\mathbf{p}_{i} = [p_{i1}, \cdots, p_{im}]^{T}$ is the $i$-th row of $\mathbf{P}$, where $p_{ij}$ denotes the chance that the node $i$ is the $j$-th important node among all $m$ nodes.
As an example, given $\mathbf{p}_{i} = [0.1, 0.8, 0.1]$, we can consider the node $i$ as the second most important node because $p_{i2}=0.8$.

For the purpose of obtaining the node confidence suggesting the importance of a vertex, we first convert the learned $\hat{\mathbf{A}}$ to the probabilistic permutation matrix $\mathbf{P}$.
Particularly, the $i$-th row of $\mathbf{P}$ is computed by
\begin{equation}\label{neuralsort}
\mathbf{p}_{i} (\mathbf{a}_{s}, \tau)= \text{Softmax} \left[\left((m+1-2i)\mathbf{a}_{s} - \Delta \mathbf{1}\right) / \tau\right],
\end{equation}
where $\tau > 0$ is a temperature parameter and $\Delta \in \mathbb{R}^{m \times m}$ is a pairwise difference matrix calculated by
\begin{equation}\label{Delta}
[\Delta]_{ij} = \left|[\mathbf{a}_{s}]_{i} - [\mathbf{a}_{s}]_{j} \right|.
\end{equation}
Vector $\mathbf{a}_{s} \in \mathbb{R}^{m}$ denotes the mean of nonzero entries in each column of the matrix $\hat{\mathbf{A}}$, measuring the average indegree of a vertex.
Namely, 
\begin{equation}
  [\mathbf{a}_{s}]_{j} = \frac{\sum_{i=1}^{m} [\hat{\mathbf{A}}]_{ij}}{\delta},
\end{equation}
where $\delta$ is the number of nonzero entries in each column.
Therefore,
Equation \eqref{neuralsort} is differentiable everywhere, which ensures the gradient computation and the back propagation during network training.
It was proved in \cite{GroverWZE19} that $\mathbf{P}$ is a unimodal row stochastic matrix where the sum of each row equals one, and the permutation can be obtained by a row-wise $\arg \max$ operator.
Note that entries in $\mathbf{P}$ stand for the chances of permutations, which can promote the flexibility of the model and avoid the negative effect of noises to some extent.
Compared with binary permutation matrices or a direct top-$k$ operator on entries of adjacency matrices, a probabilistic continuous permutation matrix enables more flexible optimization of graph-structural data with specific tasks.
Therefore, the permutation matrix $\mathbf{P}$ may provide better robustness and generalization during training iterations.

Next, we introduce a metric to evaluate the confidence of each node with the probabilistic permutation matrix.
It is notable that the probability distribution of the node $i$ should be higher in the front of vector $\mathbf{p}_{i}$, if the node $i$ has a higher chance corresponding to a larger confidence value.
Inspired by the Discounted Cumulative Gain (DCG) that is commonly employed in recommender systems and search algorithms, we calculate the confidence of the node $i$ by
\begin{equation}\label{DCG}
  \mathcal{I}_{i} = \sum_{j=1}^{m} \frac{2^{p_{ij}} - 1}{\mathrm{log}_{2} (j + 1)},
\end{equation}
and renormalize it with
\begin{equation}\label{NDCG}
  \bar{\mathcal{I}}_{i} = \frac{\mathcal{I}_{i} - \mathcal{I}_{min}}{\mathcal{I}_{max} - \mathcal{I}_{min}},
\end{equation}
where $\mathcal{I}_{max}$ and $\mathcal{I}_{min}$ are the largest and smallest values of node confidence computed by Equation \eqref{DCG}, respectively.
As a matter of fact, Equations \eqref{DCG} and \eqref{NDCG} transfer the density near each node into a probability value ranging in $[0,1]$.
Theoretically, a higher $\bar{\mathcal{I}}_{i}$ indicates a higher confidence of the node $i$, because a larger $p_{ij}$ with smaller $i$ contributes more to the accumulation computations in Equation \eqref{DCG}.
On the basis of this, the differentiable node selection procedure is performed via a coefficient matrix $\mathbf{C} \in \mathbb{R}^{m \times m}$, where each entry is computed by
\begin{gather}
  [\mathbf{C}]_{ij} = \frac{\bar{\mathcal{I}}_{i} + \bar{\mathcal{I}}_{j}}{2}.
\end{gather}
Consequently, we develop a learnable node selection schema according to previous analyses, formulated as
\begin{gather}
  \mathbf{C}_{select} = \text{ReLU} \left( \mathbf{C} - \theta \right),\label{DNodeSelection1} \\
  \mathbf{A}_{select} = \hat{\mathbf{A}} \odot \frac{\mathbf{C}_{select}}{\mathrm{max} \left(\mathbf{C}_{select} \right)}, \label{DNodeSelection2}
\end{gather}
where $\theta > 0$ serves as a learnable thresholder and $\mathrm{max} \left(\mathbf{C}_{select} \right)$ is the largest value of $\mathbf{C}_{select}$.
To ensure that $0 \leq \theta \leq 1$ holds, we employ a sigmoid projection of $\theta$ before the computation, that is, $\theta = \text{Sigmoid} \left( \theta \right)$.
With Equation \eqref{DNodeSelection2}, the sparsity of $\mathbf{C}_{select}$ is promoted due to the zero entries generated by the $\text{ReLU}$ activation.
Compared with universally adopted fixed activation functions in neural networks, we formulate it as a trainable one.
This makes the differentiable node selection schema learn a refined sparse graph in a data-driven paradigm.
Figure \ref{DNS} summarizes the differentiable node selection procedure with a simple graph.
A new refined graph is obtained by the Hadamard product of the original adjacency matrix and a coefficient matrix generated from a permutation matrix.
Because Equations \eqref{DNodeSelection1} and \eqref{DNodeSelection2} choose confident neighbors for each specific node, they can also be regarded as a differentiable edge activation procedure to reserve the salient edges.
As an example in Figure \ref{DNS}, it can be seen that an insignificant edge connecting nodes 2 and 3 is removed by the differentiable node selection procedure.
Actually, these differentiable functions implicitly perform the non-differentiable top-$k$ operator during training, where the trainable parameter $\theta$ can be regarded as a surrogate of KNN that keeps the largest $k$ values of node connections.

\begin{figure}[!tbp]
  \centering
  \includegraphics[width=\textwidth]{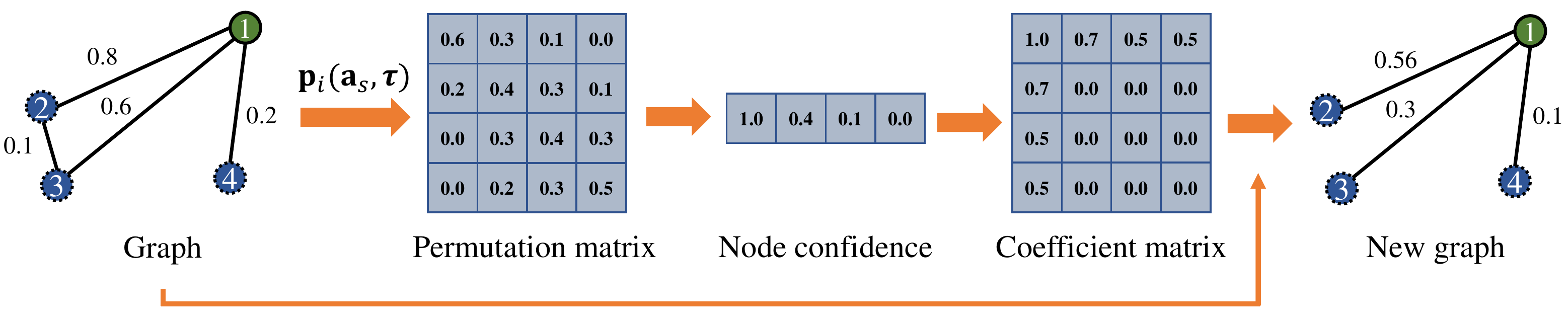}
  \caption{Differentiable node selection (edge activation) procedure. In this example,  node 2 only selects node 1 as the most important neighborhood node, resulting in the inhibition of the edge between node 2 and node 3.
  }
  \label{DNS}
\end{figure}

With the refined adjacency matrix, the proposed framework conducts graph convolution of the $l$-th layer by
\begin{gather}\label{LGCN}
\mathbf{H}^{(l)} = \sigma \left( \mathbf{\mathbf{A}}_{select} \mathbf{H}^{(l-1)} \mathbf{W}^{(l)} \right).
\end{gather}
As a straightforward example, the node embedding $\mathbf{Z}$ of a $2$-layer GCN is computed by
\begin{gather}\label{LGCN2Layer}
\mathbf{Z} = \mathrm{softmax} \left(  \mathbf{\mathbf{A}}_{select}  \sigma \left( \mathbf{\mathbf{A}}_{select} \mathbf{H} \mathbf{W}^{(1)} \right) \mathbf{W}^{(2)} \right),
\end{gather}
where $[\mathbf{Z}]_{ij}$ denotes the probability that the node $i$ belongs to the $j$-th class.
In terms of a semi-supervised classification task, the loss is measured by the cross-entropy function:
\begin{gather}\label{GCNLoss}
\mathcal{L} = - \sum_{i \in \Omega} \sum_{j=1}^{c} \mathbf{Y}_{ij} \mathrm{ln} \mathbf{Z}_{ij},
\end{gather}
where $\mathbf{Y} \in \mathbb{R}^{m \times c}$ is the ground truth of the dataset.
The loss values are calculated through the samples from the partially labeled sample set $\Omega$ accounting for a small portion of the entire dataset.

\begin{algorithm}[!htbp]
  \caption{Multi-view Graph Convolutional Network with Differentiable Node Selection (MGCN-DNS)}
  \renewcommand{\algorithmicrequire}{\textbf{Input:}}
  \renewcommand{\algorithmicensure}{\textbf{Output:}}
  \begin{algorithmic}\label{Algorithm}
  \REQUIRE Multi-view data $\mathcal{X} = \{\mathbf{X}^{(1)}, \cdots, \mathbf{X}^{(V)} \}$ and partially labeled matrix $\mathbf{Y} \in \mathbb{R}^{m \times c}$.
  \ENSURE Node embedding $\mathbf{Z}$.
  \STATE {Initialize trainable weights $ \{w^{(v, i)} \}^{V}_{v,i=1}$;}
  \STATE {Initialize hyperparameter $\gamma$ and temperature parameter $\tau$;}
  \STATE {Estimate adjacency matrices $\mathbf{A}^{(1)}, \cdots, \mathbf{A}^{(V)}$ via KNN;}
  \STATE {Initialize learnable matrices $\mathbf{S}_{1}, \mathbf{S}_{2}$ and thresholder $\theta$;}
  \STATE {Initialize learnable weight matrices of GCN;}
  \WHILE {not convergent}
     \STATE {Conduct learnable multi-channel graph fusion with Equations \eqref{WeightedA}, \eqref{NormalizationPai}, \eqref{ComputeAlpha}, \eqref{NormalizationAlpha} and \eqref{FusedA};}
     \STATE {Compute the refined $\hat{\mathbf{A}}$ with Equation \eqref{LearnS};}
     \STATE {Generate probabilistic permutation matrix $\mathbf{P}$ with Equations \eqref{neuralsort} and \eqref{Delta};}
     \STATE {Obtain the normalized confidence vector $\bar{\mathcal{I}}$ with Equations \eqref{DCG} and \eqref{NDCG};}
     \STATE {Conduct differentiable node selection via $\mathbf{C}_{select}$, according to Equations \eqref{DNodeSelection1} and \eqref{DNodeSelection2};}
     \STATE {Calculate $\mathbf{Z} = \mathbf{H}^{(L)}$ of GCN with Equation \eqref{LGCN};}
     \STATE {Evaluate the loss of the framework with Equation \eqref{GCNLoss};}
     \STATE {Update learnable parameters $\{ \mathbf{S}_{1}\}_{v=1}^{V}$, $\{ \mathbf{S}_{2}\}_{v=1}^{V}$, $\{\theta\}_{v=1}^{V}$, $\{ \mathbf{W}^{(l)} \}_{l=1}^{L}$ and $\{w^{(i, j)}\}_{i, j=1}^{V}$ with back propagation;}
  \ENDWHILE
  \RETURN {Node embedding $\mathbf{Z}$.}
  \end{algorithmic}
  \end{algorithm}

\subsection{Computational Complexity}

Algorithm \ref{Algorithm} provides an elaborate description of the proposed MGCN-DNS.
In general, the forward propagation of MGCN-DNS framework can be divided into the following steps: adaptive graph fusion, graph learning, differentiable node selection, and graph convolutions.
The graph fusion process consumes $\mathcal{O}(mn)$ and the graph learning module costs $\mathcal{O}(m^2 + m^3)$.
The computational complexity of sorting operator and node confidence computations is $\mathcal{O}(m^2)$.
Although the classical sorting algorithm only consumes an overall complexity of $\mathcal{O}(m \log m)$, the proposed node selection calculation can run efficiently on GPU hardware parallelly.
In addition, given $d$ hidden units and $d \ll n$, the computational consumption of graph convolution is $\mathcal{O}(mnd)$.
Consequently, the upper bound on time consumption of forward calculations in an iteration is $\mathcal{O} \left( m (m^{2} + nd + m + n)   \right)$.

\section{Experimental Results}\label{experiments}

In this section, the proposed MGCN-DNS is evaluated by substantial experiments to verify its effectiveness.
Several benchmark datasets and state-of-the-art methods are selected to assess the performance of MGCN-DNS.
The framework is implemented with Pytorch platform and run on the computer with AMD R9-5900X CPU, RTX 3060 GPU and 32G RAM.
\subsection{Dataset Descriptions}
Six publicly available multi-view datasets are selected for performance comparison:
Caltech-20, Caltech-all\footnote{http://www.vision.caltech.edu/Image\_Datasets/Caltech101/Caltech101.html}, BBCnews\footnote{http://mlg.ucd.ie/datasets/segment.html}, BBCsports\footnote{http://mlg.ucd.ie/datasets/bbc.html}, MNIST, MNIST-10K\footnote{http://yann.lecun.com/exdb/mnist/}, NUS-WIDE\footnote{https://lms.comp.nus.edu.sg/wp-content/uploads/2019/research/nuswide/NUS-WIDE.html} and Youtube\footnote{http://archive.ics.uci.edu/ml/machine-learning-databases/00269/}, which include multiple descriptions of images or documents.
The features of each view in these datasets are extracted by some well-known feature extraction algorithms.
Here we provide a more detailed introduction to these datasets:

(1) \textbf{Caltech-20} and \textbf{Caltech-all}: Caltech101 is a well-known object recognition dataset including 101 classes of images.
Six multifarious features are extracted: 48-dimensional Gabor, 40-dimensional wavelet moments, 254-dimensional CENTRIST, 1,984-dimensional histogram of oriented gradients, 512-dimensional GIST and 928-dimensional LBP features.
We follow prior work \cite{NieLiLi17SelfWeighted} where 20 classes are selected to generate the Caltech-20 subset.

(2) \textbf{BBCnews}: It is a collection of news reports which covers politics, entertainment, business, sport and technology fields.  The dataset is constituted of 4 views via splitting documents into 4 related segments.

(3) \textbf{BBCsports}: Different from BBCnews, it is a dataset consisting of diverse areas from BBC sports websites, including football, athletics, cricket, rugby and tennis news, which are illustrated from 2 distinct views.

(4) \textbf{MNIST} and \textbf{MNIST-10K}: It is a well-known dataset of handwritten digits, where 3 types of features are extracted: 30-dimensional IsoProjection, 9-dimensional linear discriminant analysis features and 9-dimensional neighborhood preserving embedding features.

(5) \textbf{NUS-WIDE}: This is a popular dataset for object recognition tasks. We randomly choose 8 classes and 6 feature sets: 64-dimensional color histogram, 144-dimensional color correlogram, 73-dimensional edge direction histogram, 128-dimensional wavelet texture, 225-dimensional block-wise color moments and 500-dimensional bag of words from SIFT descriptors.

(6) \textbf{Youtube}: This dataset contains various audio and visual features on video data, including 2,000-dimensional mfcc, 64-dimensional volume stream, 1,024-dimensional spectrogram stream, 512-dimensional cuboids histogram, 64-dimensional hist motion estimate and 647-dimensional HOG features.

A summary of all datasets is presented in Table \ref{DataDescription}, including statistics of numbers of samples, views, classes, features and data types.

\begin{table*}[!tbp]
  \center
  \begin{tabular}{c|ccccc}
  \toprule
   Datasets&      \# Samples & \# Views   & \# Features               & \# Classes & Data Types\\
   \midrule
   Caltech-20      & 1,474    & 6          & 48/40/254/1,984/512/928   & 7          & Object images              \\
   Caltech-all     & 9,144    & 6          & 48/40/254/1,984/512/928   & 20         & Object images       \\
   BBCnews        & 685      & 4          & 4,659/4,633/4,665/4,684   & 5          & Documents \\
   BBCsports      & 544      & 2          & 3,183/3,203               & 5          & Documents           \\
   MNIST          & 2,000    & 3          & 30/9/9                    & 10         & Digit images           \\
   MNIST-10K      & 10,000    & 3          & 30/9/9                    & 10         & Digit images           \\
   NUS-WIDE       & 1,600    & 6          & 64/144/73/128/225/500     & 8          & Web images            \\
   Youtube      & 2,000      & 6          & 2,000/64/1,024/512/64/647                    & 10         & Video data            \\
  \bottomrule
  \end{tabular}
  \caption{A brief statistics of all test multi-view datasets.}
  \label{DataDescription}
  \end{table*}


\subsection{Compared Methods}
So as to validate the effectiveness of the proposed MGCN-DNS, we compare the performance of semi-supervised classification tasks with several traditional and state-of-the-art methods, including KNN, SVM, AMGL \cite{NieLiLi16Parameter}, MVAR \cite{TaoHNZY17}, MLAN \cite{NieCaiLi17Multiview}, AWDR \cite{YangDN19}, HLR-$\mathbf{M}^2$VS \cite{XieZQDT20}, ELR-MVSC \cite{HuangWZZL21}, DSRL \cite{WangCDL22}, GCN \cite{KipfW17}, Co-GCN \cite{LiLW20a} and SSGCN \cite{ZhuK21}. 
For the sake of reproducibility and providing a fair test bed, the configurations are set empirically during experiments or defined by default settings as original papers.

It is noteworthy that there are only three GCN-based baselines (GCN, Co-GCN and SSGCN), attributed to the fact that limited work has focused on GCN conducting downstream classification tasks with multi-view data.
Because GCN and SSGCN can only manipulate single-view data, we adopt average weighted adjacency matrices as inputs.
Nevertheless, most selected approaches are graph-based models so that we are able to conduct a convincing comparison.

\subsection{Evaluation Metrics}
The performance of each individual algorithm is evaluated via classification accuracy, that is, the percentage of samples that are correctly classified.
Following some prior works \cite{LiLW20a,XieZQDT20,HuangWZZL21}, we conduct semi-supervised classification experiments by splitting data into the training set and testing set, where the number of training samples is smaller than the testing samples.
We terminate all gradient-based methods including the proposed MGCN-DNS when they reach the maximum iterations or their losses converge.
All approaches are executed 5 times with randomly selected labeled samples at each time.
Accordingly, we record the average classification accuracy and its standard deviation in all comparative experiments.

\begin{table*}[!tbp]
  \centering
  \resizebox{\textwidth}{!}{
  \begin{tabular}{c|c|cccccccc}
  \toprule
  Ratios & Methods $\backslash$ Datasets        & Caltech-20         & Caltech-all            & BBCnews                & BBCsports              & MNIST         & MNIST-10K  & NUS-WIDE  & Youtube           \\
  \midrule 
  \multirow{13}{*}{$1\%$} & KNN   & 51.2 (0.4)        & 19.5 (1.9)            & 26.9 (9.7)             & 28.1 (9.6)             & 62.4 (7.4)  & 74.8 (1.0)  &  23.6 (3.1) & 7.75 (1.0)          \\
  &SVM & 53.9 (1.9)        & 20.3 (2.2)         & 32.2 (9.3)             & 29.5 (8.0)             & 63.8 (6.8)          & 77.3 (2.6)  & 26.3 (1.6)  & 21.0 (6.1)   \\ \cmidrule{2-10}
  &AMGL \cite{NieLiLi16Parameter} & 27.2 (2.7)        & 11.2 (3.1)         & 32.5 (3.4)             & 35.1 (4.3)             & 61.5 (4.1)          & 85.9 (5.2)  & 29.9 (4.3)  & 5.31 (0.9)   \\
  &MVAR \cite{TaoHNZY17} & 45.8 (9.6)        & 23.0 (0.2)            & 28.1 (6.5)             & 35.5 (5.9)             & 49.7 (7.6)     & 82.3 (0.1)        & 15.3 (2.7)  & 4.72 (1.3)   \\
  &MLAN \cite{NieCaiLi17Multiview} & 21.2 (2.4)        & 11.5 (2.2)            & 29.6 (8.1)             & 41.0 (7.9)             & 57.5 (5.2)     & 85.6 (3.5)        & 23.0 (9.1)  & 2.50 (0.2)   \\
  &AWDR \cite{YangDN19} & 47.7 (6.5)        & 16.5 (0.4)            & 32.2 (9.9)             & 20.0 (5.1)             & 50.1 (4.1)     & 66.8 (2.3)      & 22.3 (4.0)  & 31.6 (8.5)   \\
  &HLR-$\mathrm{M}^2$VS \cite{XieZQDT20}&  60.7 (9.3)        & -            & 50.0 (9.3)             & 40.9 (9.1)             & 63.3 (6.7)   & -       & 18.8 (3.6)  &  8.03 (1.1)       \\ 
  &ERL-MVSC \cite{HuangWZZL21}&  57.2 (1.9)        & 21.5 (0.4)            & 57.7 (4.0)             & 66.7 (7.5)             & 77.6 (4.8)   & 83.4 (0.3)       & 22.0 (2.2)  &  26.9 (2.1)       \\
  &DSRL \cite{WangCDL22}& 65.5 (2.6)  & 22.5 (0.6) & 59.4 (9.7)  &  66.5 (9.5)  & 67.1 (8.2) &  72.8 (7.3)  & 31.4 (2.9) & 26.7 (1.7) \\  \cmidrule{2-10}
  &GCN fusion \cite{KipfW17} & 66.0 (1.2)        & 24.4 (0.7)            & 72.9 (3.0)             & 81.2 (1.1)             & 80.6 (3.1)     & 81.8 (1.2)        & \textcolor{blue}{\textbf{34.2 (2.4)}}  & 29.5 (2.5)  \\
  &Co-GCN  \cite{LiLW20a}    & \textcolor{red}{\textbf{70.3 (1.2)}}        & 25.1 (0.5)            & 73.0 (1.0)             & \textcolor{blue}{\textbf{81.4 (9.1)}}             & 80.7 (0.7)     & 83.3 (0.9)    & 32.0 (2.3)  &  \textcolor{blue}{\textbf{31.2 (3.1)}}   \\
  &SSGCN  \cite{ZhuK21}     & 66.4 (2.4)        & \textcolor{blue}{\textbf{25.8 (0.8)}}            & \textcolor{blue}{\textbf{74.9 (3.3)}}             & 80.4 (2.1)            & \textcolor{blue}{\textbf{82.9 (3.4)}}    &  \textcolor{blue}{\textbf{83.9 (0.7)}}    & 33.9 (1.3)  & 28.9 (4.3)    \\ \cmidrule{2-10}
  &MGCN-DNS    & \textcolor{blue}{\textbf{69.3 (1.3)}}& \textcolor{red}{\textbf{26.2 (1.4)}}  & \textcolor{red}{\textbf{77.8 (5.4)}}    & \textcolor{red}{\textbf{83.5 (9.6)}}    & \textcolor{red}{\textbf{88.0 (1.3)}}  & \textcolor{red}{\textbf{88.5 (0.8)}}  & \textcolor{red}{\textbf{44.2 (3.9)}}  & \textcolor{red}{\textbf{33.4 (6.6)}}   \\
  \midrule \midrule
  \multirow{13}{*}{$10\%$} & KNN   & 67.0 (0.4)        & 29.6 (0.3)            & 38.3 (9.6)             & 38.3 (9.5)             & 82.4 (0.9)  & 89.8 (0.4)  &  32.7 (2.9) & 36.8 (1.6)          \\
  &SVM & 71.1 (1.0)        & 35.3 (0.4)         & 72.9 (7.0)             & 70.9 (9.1)             & 86.2 (0.9)          & 88.6 (0.5)  & 43.1 (1.4)  & 45.6 (1.0)   \\ \cmidrule{2-10}
  &AMGL \cite{NieLiLi16Parameter} & 45.0 (3.1)        & 24.4 (0.2)            & 52.3 (5.5)             & 55.6 (1.4)             & 69.5 (0.8)          & 88.5 (0.2)  & 41.9 (0.2)  & 44.2 (0.8)   \\
  &MVAR \cite{TaoHNZY17} & 68.9 (4.6)        & 43.5 (0.3)            & 75.3 (5.5)             & 83.7 (3.8)             & 83.4 (1.7)     & 85.3 (0.8)        & 33.3 (2.6)  & 37.0 (1.3)   \\
  &MLAN \cite{NieCaiLi17Multiview} & 45.9 (1.7)        & 31.2 (0.5)            & 74.1 (0.9)             & 62.6 (2.2)             & 69.8 (1.4)     & 88.6 (0.3)        & 27.0 (1.4)  & 36.4 (1.0)   \\
  &AWDR \cite{YangDN19} & 49.2 (1.7)        & 45.0 (1.3)            & 85.7 (1.4)             & 81.3 (3.3)             & 84.0 (2.6)     & 78.1 (0.3)      & 40.6 (1.9)  & 36.1 (2.8)   \\
  &HLR-$\mathrm{M}^2$VS \cite{XieZQDT20}&  80.5 (1.8)        & -            & 78.1 (2.8)             & 84.6 (0.4)             & 87.1 (0.4)   & -       & 35.3 (1.7)  &  51.2 (4.1)       \\ 
  &ERL-MVSC \cite{HuangWZZL21}&  83.2 (1.6)       & 46.1 (0.3)            & 85.9 (2.2)             & 90.3 (1.9)             & 87.2 (0.2)   & 90.9 (0.3)       & \textcolor{blue}{\textbf{44.4 (2.1)}}  &  57.4 (1.8)       \\ 
  &DSRL \cite{WangCDL22}&  \textcolor{blue}{\textbf{84.1 (2.0)}}   &  \textcolor{blue}{\textbf{51.1 (0.2)}}  &  88.5 (1.2) & 91.6 (5.4)   & 87.6 (1.2) & 89.3 (0.4) &  44.3 (2.4) & 48.0 (1.4)   \\  \cmidrule{2-10}
  &GCN fusion \cite{KipfW17} & 79.9 (0.3)        & 44.2 (0.3)            & 69.6 (1.8)             & 87.0 (2.0)             & 85.0 (0.7)     & 89.3 (0.5)        & 42.9 (2.1)  & 57.8 (1.4)  \\
  &Co-GCN  \cite{LiLW20a}    & 78.2 (0.8)        & 46.2 (0.6)            & 81.9 (1.5)             & 84.8 (1.4)             & 87.9 (0.5)     & \textcolor{blue}{\textbf{91.2 (0.4)}}    & 43.5 (0.4)  &  \textcolor{blue}{\textbf{58.0 (0.5)}}   \\
  &SSGCN  \cite{ZhuK21}     &82.5 (0.7)        & 47.1 (0.1)           & \textcolor{blue}{\textbf{90.7 (0.5)}}             & \textcolor{blue}{\textbf{94.1 (0.1)}}             & \textcolor{blue}{\textbf{88.4 (0.2)}}     & 89.9 (0.1)    & 41.2 (0.4)  &  57.1 (1.1)   \\ \cmidrule{2-10}
  &MGCN-DNS    & \textcolor{red}{\textbf{84.3 (1.2)}} & \textcolor{red}{\textbf{53.3 (0.3)}}  & \textcolor{red}{\textbf{93.6 (0.6)}}    & \textcolor{red}{\textbf{97.4 (0.6)}}    & \textcolor{red}{\textbf{90.2 (0.6)}}  & \textcolor{red}{\textbf{92.2 (0.2)}}  & \textcolor{red}{\textbf{69.2 (3.1)}}  & \textcolor{red}{\textbf{60.5 (1.3)}}   \\
  \bottomrule
  \end{tabular}}
  \caption{Classification accuracy (mean\% and standard deviation\%) of all compared methods with $1 \% $ and $ 10 \%$  labeled samples, where the best performance is highlighted in red and the second-best result is highlighted in blue.
  We can not obtain results of HLR-$\mathrm{M}^2$VS on some datasets due to its high computational complexity, and mark the results with "-".
  }
  \label{ACCcomparsionClassification}
\end{table*}

\subsection{Experimental Results}

\subsubsection{Performance Comparison}
\begin{figure*}[!tbp]
  \centering
  \includegraphics[width=\textwidth]{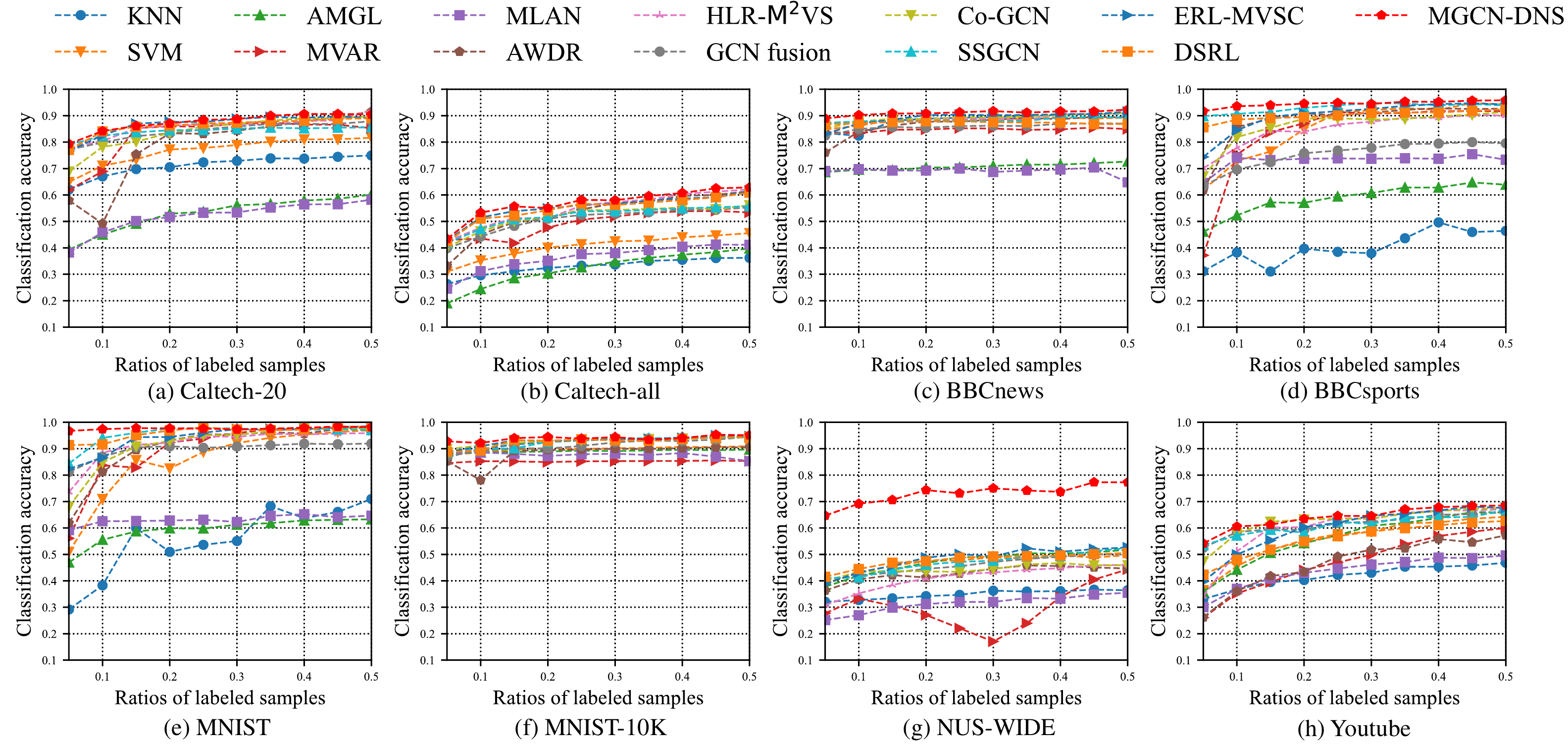}
  \caption{The varied performance (classification accuracy) of all compared methods as the ratio of labeled data ranges in $\{ 0.05, 0.10, \cdots, 0.50\}$ on all test datasets.}
  \label{SemiRatio}
\end{figure*}

First of all, we perform a substantial comparison of MGCN-DNS and other classical and state-of-the-art methods in terms of semi-supervised classification, measured by classification accuracy.
With regard to experimental settings of the proposed MGCN-DNS, the learning rate is fixed as $0.1$ and a 2-layer graph convolution structure is applied.
The updates of all trainable parameters are conducted by the Adam optimizer.
The experimental results of semi-supervised classification are reported in Table \ref{ACCcomparsionClassification}, recording the performance when $1\%$ and $10\%$ data are randomly labeled respectively.
The table reveals that the proposed MGCN-DNS achieves encouraging performance with varying ratios of labeled data, leading in all compared methods on almost all datasets.
Overall, the performance leading of GCN-based models is more remarkable when only $1\%$ samples are labeled,
and the proposed MGCN-DNS gains competitive classification accuracy in this case.
Figure \ref{SemiRatio} further provides the classification accuracy of MGCN-DNS with a larger proportion of labeled samples.
It points out that the performance leading is more considerable with a small number of labeled data (e.g. $5\%$ known labels), and the accuracy gaps among multifarious methods narrow as the ratio of labeled instances rises.
These observations indicate that MGCN-DNS succeeds in exploiting the properties of graph-structural data with extremely limited supervised information.
It is noted that MGCN-DNS also obtains pleasurable results compared with other state-of-the-art GCN-based frameworks, which verifies that the proposed model promotes the performance of GCN in terms of multi-view semi-supervised classification.

\subsubsection{Ablation Study}
We also examine the effectiveness of the graph learning module and the differentiable node selection operation by conducting ablation experiments on all datasets, as demonstrated in Table \ref{AblationStudy}.
The experimental results indicate that either of two modules succeeds in improving the performance of the entire framework, and the accuracy improvement is more favorable when both operations are conducted.
These observations validate that MGCN-DNS framework executes sustained explorations of node representation through the graph learning module and the differentiable node selection process, which facilitate improving the accuracy of downstream classification tasks.

\begin{table*}[!htbp]
  \centering
\begin{tabular}{c|ccc||c|ccc}
  \toprule
Datasets   / Modules        & GLM                       & DNS                       & Accuracy (std\%) & Datasets / Modules           & GLM                       & DNS                       & Accuracy (std\%) \\ \midrule
\multirow{4}{*}{Caltech-20} &                           &                           & 82.2 (1.2)       & \multirow{4}{*}{Caltech-all} &                           &                           & 44.2 (0.3)       \\
                            & \checkmark &                           & 83.9 (1.2)       &                              & \checkmark &                           & 46.2 (0.4)       \\
                            &                           & \checkmark & 82.8 (1.8)       &                              &                           & \checkmark & 45.7 (0.5)       \\
                            & \checkmark & \checkmark & \textbf{84.3 (1.2)}       &                              & \checkmark & \checkmark & \textbf{48.6 (0.3)}       \\ \midrule
\multirow{4}{*}{MNIST}      &                           &                           & 86.9 (0.2)       & \multirow{4}{*}{MNIST-10K}    &                           &                           & 90.3 (0.5)       \\
                            & \checkmark &                           & 88.9 (0.3)       &                              & \checkmark &                           & 90.5 (0.8)       \\
                            &                           & \checkmark & 88.8 (0.4)       &                              &                           & \checkmark & 91.1 (0.3)       \\
                            & \checkmark & \checkmark & \textbf{90.2 (0.6)}       &                              & \checkmark & \checkmark & \textbf{92.2 (0.2)}       \\ \midrule
\multirow{4}{*}{BBCnews}    &                           &                           & 90.8 (1.2)       & \multirow{4}{*}{BBCsports}   &                           &                           & 95.6 (0.5)       \\
                            & \checkmark &                           & 92.9 (1.0)       &                              & \checkmark &                           & 96.8 (0.6)       \\
                            &                           & \checkmark & 93.2 (0.5)       &                              &                           & \checkmark & 96.2 (0.9)      \\
                            & \checkmark & \checkmark & \textbf{93.6 (0.6)}       &                              & \checkmark & \checkmark & \textbf{97.4 (0.6)}       \\ \midrule
\multirow{4}{*}{NUS-WIDE}   &                           &                           & 61.9 (3.9)       & \multirow{4}{*}{Youtube}     &                           &                           & 57.5 (0.9)       \\
                            & \checkmark &                           & 65.3 (2.4)       &                              & \checkmark &                           & 59.5 (0.9)       \\
                            &                           & \checkmark & 66.1 (2.9)       &                              &                           & \checkmark & 59.0 (0.3)       \\
                            & \checkmark & \checkmark & \textbf{69.2 (3.1)}       &                              & \checkmark & \checkmark & \textbf{60.5 (1.3)}      \\ 
                            \bottomrule
\end{tabular}
\caption{Ablation study (mean classification accuracy (std\%)) of MGCN-DNS on all datasets. GLM: Graph Learning Module. DNS: Differentiable Node Selection.}
\label{AblationStudy}
\end{table*}

\begin{figure*}[!htbp]
  \centering
  \includegraphics[width=\textwidth]{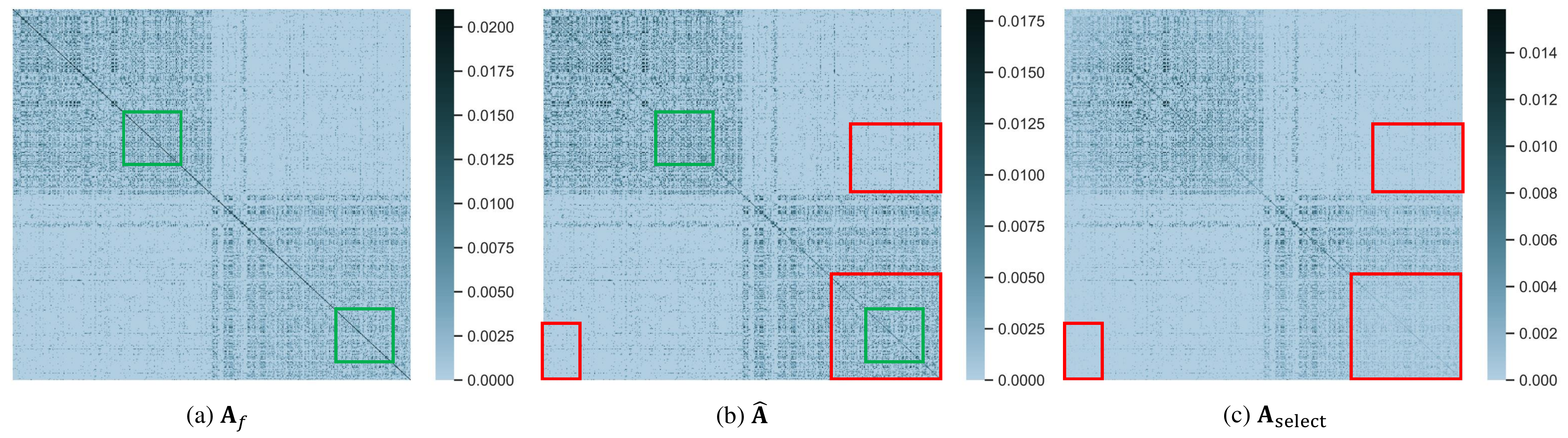}
  \caption{Visualization of partial adjacency matrices $\mathbf{A}_{f}$, $\hat{\mathbf{A}}$ and $\mathbf{A}_{select}$ on NUS-WIDE dataset.}
  \label{AdjVisualization}
\end{figure*}

In addition, we visualize partial learned adjacency matrices $\mathbf{A}_{f}$, $\hat{\mathbf{A}}$ and $\mathbf{A}_{select}$ in the network, as exhibited in Figure \ref{AdjVisualization}.
These figures suggest that $\hat{\mathbf{A}}$ learned by the graph learning module is distinct from the automatically fused adjacency matrix $\mathbf{A}_{f}$, which inhibits some node relationships owing to the adaptive shrinkage weight matrix (e.g., green boxes in Figure \ref{AdjVisualization}).
Nevertheless, most distinctive connections are retained, and the refined adjacency matrix is still symmetric.
Due to the proposed differentiable node selection operator, most salient edges are highlighted and some connective information is not selected in $\mathbf{A}_{select}$, as shown in the red boxes in Figure \ref{AdjVisualization}.
Consequently, $\mathbf{A}_{select}$ is sparser than previous adjacency matrices.
In summary, the ablation study validates that the learned sparse $\mathbf{A}_{select}$ is beneficial to the information propagation over the graph and the performance improvement of GCN.

\subsubsection{Parameter Sensitivity}
\begin{figure}[!tbp]
  \centering
  \includegraphics[width=0.7\textwidth]{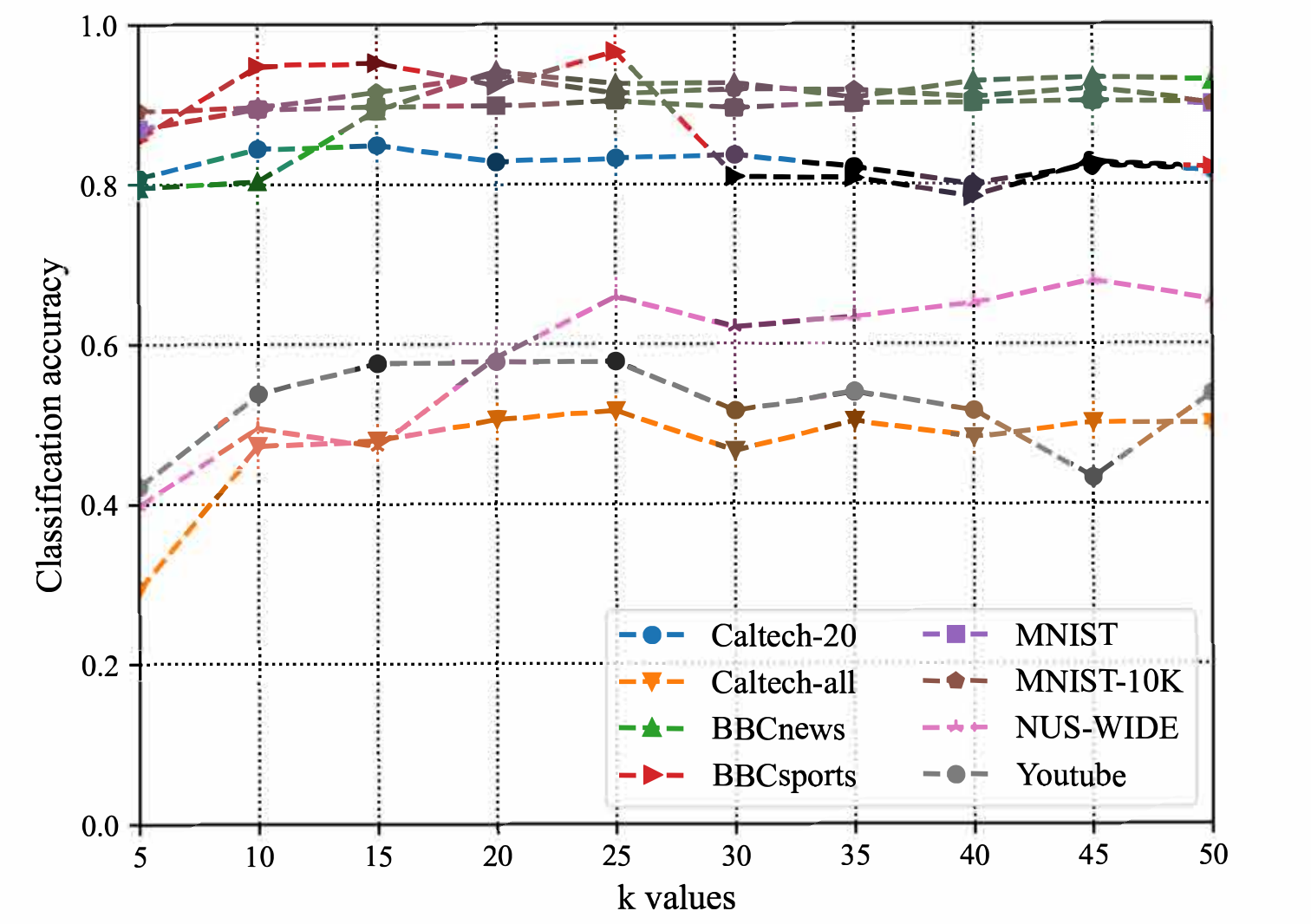}
  \caption{Parameter sensitivity (Accuracy) w.r.t. neighbor numbers $k$ on all test datasets.}
  \label{Parak}
\end{figure}

First, we analyze the effect of neighbor numbers $k$ when constructing multi-view graphs, shown in Figure \ref{Parak}.
The experimental analysis points out that the performance of MGCN-DNS is sensitive to the initialization of adjacency matrices, which is tightly relevant to the neighbor number $k$.
Overall, a small $k$ (e.g., $k=5$) often results in poor classification performance, and the accuracy rises as the number of neighbors increases.
Nonetheless, on some datasets like BBCsports and Youtube, constructing node relationships with excessive neighbors lead to significant performance decline.
This may be due to the fact that too many neighbors may introduce unexpected noises.
In conclusion, the optimal values of $k$ for constructing adjacency matrices are quite different on various datasets.
Because $k$ is directly related to the performance of graph-based models, we ought to set a suitable $k$ for each dataset when establishing topological networks.

\begin{figure*}[!tbp]
  \centering
  \includegraphics[width=\textwidth]{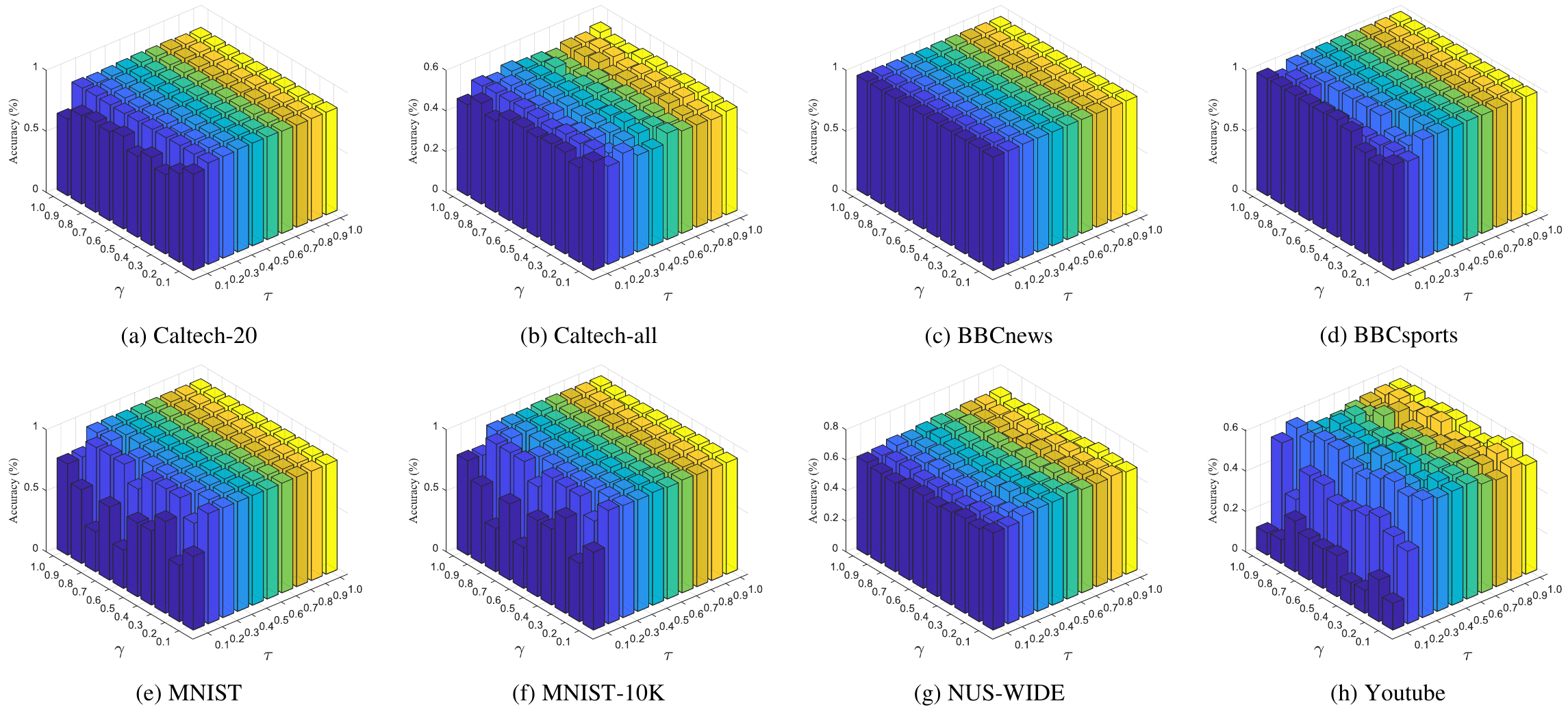}
  \caption{Parameter sensitivity (Accuracy) w.r.t. hyperparameters $\gamma$ and $\tau$ on all test datasets.}
  \label{ParaSen}
\end{figure*}

Furthermore, Figure \ref{ParaSen} explores the effect of hyperparameters $\gamma$ and $\tau$ on selected datasets.
From this figure, we can draw the following conclusions.
It can be observed that the selection of these two hyperparameters has a noticeable influence on classification performance.
In general, small values of $\gamma$ and $\tau$ often lead to poor accuracy on most datasets, which is especially significant for $\tau$.
On all datasets except BBCnews, MGCN-DNS behaves unsatisfactorily when $\tau < 0.3$.
The reason for the unfavorable performance may be that the permutation matrix $\mathbf{P}$ becomes an exact one rather than an approximate and continuous one, when a small $\tau$ is adopted.
In most cases, a probabilistic and continuous permutation matrix is helpful to the dexterous optimization of the model.
In general, the classification accuracy is robust to the change of hyperparameters when $\gamma$ and $\tau$ are large enough on most datasets.
On Youtube dataset, the accuracy fluctuates marginally when $\tau > 0.4$, indicating that choosing parameters tailored for data is critical on some datasets.

\begin{figure*}[!tbp]
  \centering
  \includegraphics[width=0.8\textwidth]{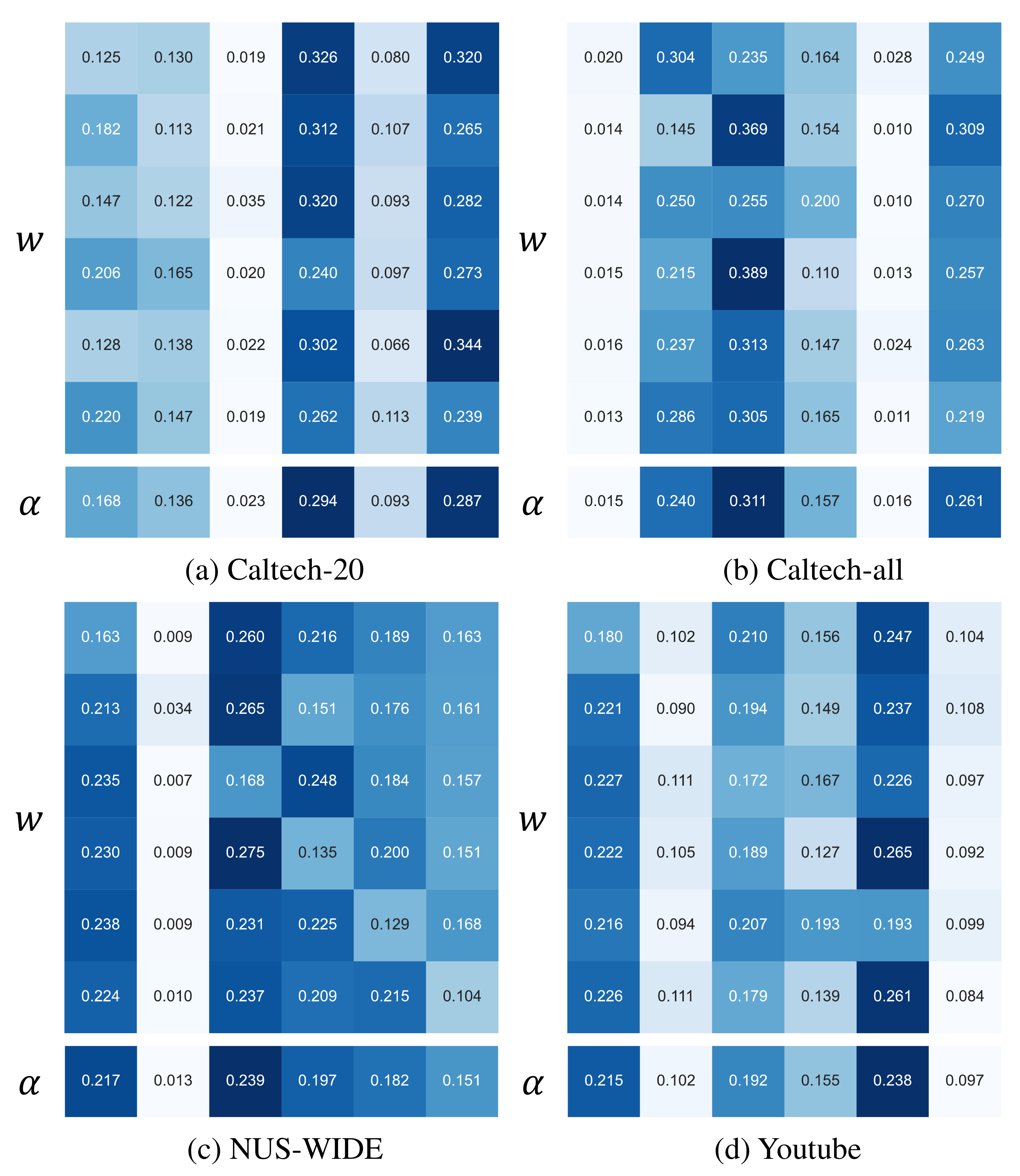}
  \caption{Visualization of trainable weights $w$ and $\alpha$ on Caltech-20, Caltech-all, NUS-WIDE and Youtube datasets.
  }
  \label{weights}
\end{figure*}

\subsubsection{Visualization of Weights}
We visualize the learnable weights $\{ w^{(i,j)} \}_{i,j=1}^{V}$ and $\{\alpha^{(i)} \}_{i=1}^{V}$ in Figure \ref{weights}.
It can be validated that KNN graphs from some views are essential for providing complementary information, attributed to which these views are more important.
Accordingly, MGCN-DNS adaptively assigns higher $\alpha$ values to these views for the sake of fully leveraging complementary information.
The model also assigns tiny values to some views, which means that these views have little complementary information.
Besides, a direct connection of initial KNN may be problematic because varying types of features in multi-view data have assorted meanings. 
This may result in the dissonance of information from different views.
Thus, we develop a two-stage fusion process which generates the combined $\mathbf{A}_{f}$ via a linear weighted accumulation on the basis of learned $\{\alpha^{(i)} \}_{i=1}^{V}$.

\subsubsection{Training Details}
\begin{figure*}[!tbp]
  \centering
  \includegraphics[width=\textwidth]{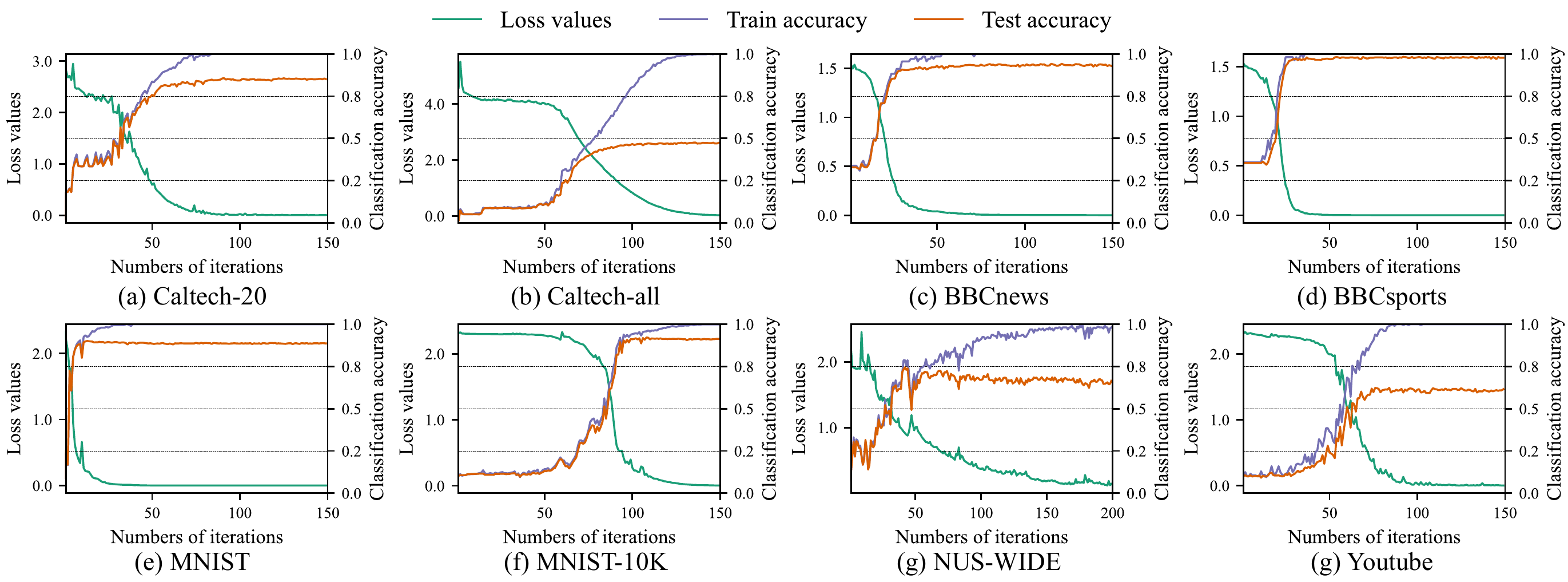}
  \caption{Curves of loss values (green), train accuracy (purple) and test accuracy (orange) with $10 \%$ labeled samples on all datasets.
  }
  \label{convergence}
\end{figure*}

Furthermore, we analyze the convergence of the proposed framework, and look into the classification accuracy of train sets and test sets during the training procedure, recorded in Figure \ref{convergence}.
The experimental results demonstrate that the loss values of MGCN-DNS plunge and converge rapidly, swelling to the lowest point within 200 iterations, suggesting the high efficiency of the framework.
The classification accuracy of the train set soars and stabilizes at a peak during training.
Although the predictive accuracy of unlabeled data shoots up as training continues, it reaches a lower plateau compared with training accuracy, and begins to fluctuate or even dwindle when the training is not finished, indicating that the phenomenon of overfitting may take place with an excessive number of iterations.

\section{Conclusion}\label{conclusion}
In this work, we proposed a multi-channel GCN-based model with a differentiable node selection schema.
The proposed method applied multi-graph-structural data for initialization and aimed to learn topological information with better robustness and generalization.
In each channel, different initialized graphs were integrated by a two-stage adaptive graph fusion procedure.
The fused graph was refined by a graph learning module and the differentiable operation of node selection, which explored graph embeddings with higher quality to promote the performance of graph convolutions.
Eventually, the proposed framework was applied to carrying out multi-view semi-supervised classification tasks.
Comprehensive experimental results validated the effectiveness of MGCN-DNS and revealed that it gained encouraging performance improvement.

In the future, we will devote ourselves to further study on multi-channel GCN-based models and exploit insightful models to conduct sustained explorations of graph embeddings with better robustness and explainable semantic information.
A more dexterous strategy to fuse graphs from heterogeneous sources is also our next direction.

\section{Acknowledgments}
This work is in part supported by the National Natural Science Foundation of China (Grant Nos. U21A20472 and 62276065), the Natural Science Foundation of Fujian Province (Grant No. 2020J01130193).

\bibliographystyle{ACM-Reference-Format}
\bibliography{MachineLearning}
\end{document}